# Towards AI enabled automated tracking of multiple boxers

A.S. Karthikeyan, Vipul Baghel, Anish Monsley Kirupakaran, John Warburton, Ranganathan Srinivasan, Babji Srinivasan*, Ravi Sadananda Hegde**

*Abstract*—Continuous tracking of boxers across multiple training sessions helps quantify traits required for the well-known ten-point-must system. However, continuous tracking of multiple athletes across multiple training sessions remains a challenge, because it is difficult to precisely segment bout boundaries in a recorded video stream. Furthermore, re-identification of the same athlete over different period or even within the same bout remains a challenge. Difficulties are further compounded when a single fixed view video is captured in top-view. This work summarizes our progress in creating a system in an economically single fixed top-view camera. Specifically, we describe improved algorithm for bout transition detection and in-bout continuous player identification without erroneous ID updation or ID switching. From our custom collected data of ~11 hours (athlete count: 45, bouts: 189), our transition detection algorithm achieves 90% accuracy and continuous ID tracking achieves IDU=0, IDS=0.

## I. INTRODUCTION

Starting from Rio Olympics, the international boxing association (AIBA) and international Olympic committee (IOC) adopted the ten-point-must system rule for Olympic boxing in order to improve the consistency. These rules emphasize certain traits like quantity and quality of punches, dominance, and competitiveness of boxers. It will be immensely beneficial to continuously track these traits of multiple boxers across multiple training sessions. Therein, an vision based system can be helpful in reducing the burden of continuous tracking. But because boxing is a dynamic sport, most video capture systems usually require multiple cameras at multiple viewpoints (overhead, side, top, and close-up view) which significantly increases the cost. Thus, there lies a deep interest in automated tracking of boxers through computer vision techniques.

Our work is based on wide spread adaption of automated systems using single camera. Moreover, video transition system and person re-identification are well known tasks in computer vision. Hence, we specifically report our progress in the top-view. For the ease of installation, we chose fixed top-view which provide an unconventional view. In order to perform analytics, we need precisely cut training sessions into well-defined bouts, and perform in-bout tracking in these segmented bouts. However, some of the major challenges that exist in top-view data are transitions are hard to spot, boxers in a bout wearing the same attire, loss of anthropometric features, etc. In addition to that, maintaining ID continuity within a same bout is difficult, especially when they are very close to each other. Any inconsistency in tracking may lead to erroneous ID updation (IDU), and ID Switching (IDS)

Literature suggests that works carried out in transition detection have used side or multimodal information. Researchers have used various techniques like pixel differences, motion information, machine learning, and mixed method [1-3]. For person re-identification, most researchers have made use of side view data combined with machine learning techniques [4] or commercial software [5], to detect and track these individuals. While, few researchers have utilized depth sensors to work on top-view data [6]. With the introduction of frameworks like mediapose, alphasepose, openpose, and DECA, estimating landmarks has become robust [7-8]. However, all these models have been developed with purview of side view. Pose estimation can be performed through top-down or bottom-up approaches. The former approach detects the individual followed by estimation of landmarks while the later approach does the converse. Also, the usability or its performance from top-view data is still not investigated. Determining video transition or continuous tracking from an overhead view or top-view using RGB camera is yet to be explored. Thus, the specific contributions of this work are

- *Video segmentation into bouts*: Automatic segmentation of the training videos into bouts.
- *In-bout person re-identification*: Continuous tracking of boxers in the ring through descriptor and pose.

## II. CUSTOM VIDEO DATASET

We have collected 10 hours and 21 minutes training session videos of boxing from Inspire Institute of Technology. The boxing data has been collected from 45 athlete. These data were collected for a period of more than 4 months. The redundant recordings at the start and end of training sessions were removed manually. The total number of bouts present in this dataset is 189. The data collection setup is shown in Fig. 1 (a). Vision Alvium with frame rate of 70 FPS was used to acquire the data. To maintain the continuity of boxing rounds, any redundant recordings at the start and end of the training session were removed. Each bout was held for a duration of two minutes followed by one minute gap interval.

Research supported by Centre of Excellence of Sports Science and Analytics, Indian Institute of Technology Madras.

Anish Monsley Kirupakaran, A.S. Karthikeyan, and *Babji Srinivasan are with the Applied Mechanics Department, Indian Institute of Technology Madras, Tamil Nadu, India (ic37215@imail.iitm.ac.in; am22d006@smail.iitm.ac.in; *babji.srinivasan@iitm.ac.in).

John Warburton is with Liverpool John Moores University, England, United Kingdom. (john.warburton@inspireinstituteofsport.com)

Ranganathan Srinivasan is with Chemical Engineering Department, Indian Institute of Technology Madras, Tamil Nadu, India. (ranga@iitm.ac.in)

Vipul Baghel and **Ravi Sadananda Hegde are with Electrical Engineering Department, Indian Institute of Technology Gandhinagar, Gujarat, India. (baghelvipul@iitgn.ac.in, **hegder@iitgn.ac.in)

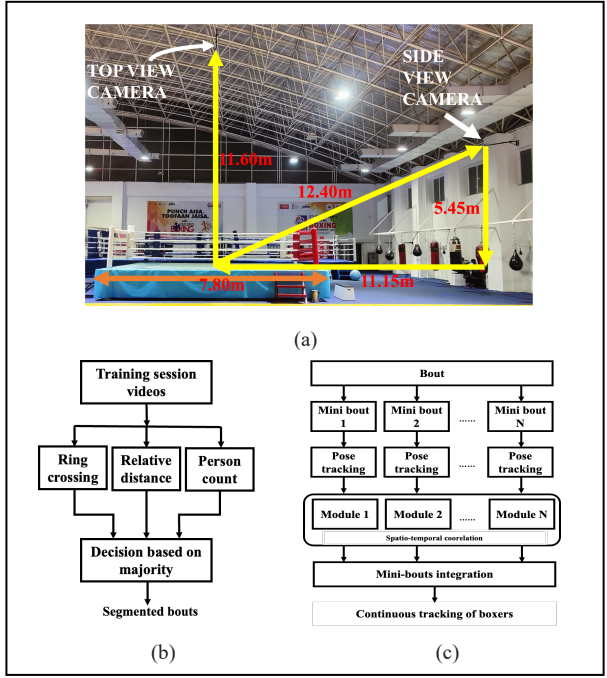

Fig. 1. Schematic diagram of boxer system - (a) custom video collection setup; (b) block diagram of automatic video transition detection; (c) block diagram of modified pose tracking

## III. PROPOSED SYSTEM

### A. Video Segmentation

The training sessions are fed as input to the proposed system, which then performs the task of (i) transition detection to segment the raw videos into bouts; (ii) in-bout boxer re-identification through detection and tracking of the boxers. Once the person is tracked throughout the video, we calculate certain metrics like hotspot and line-of-sight estimation for capturing traits.

Motivated by the observation from the input data acquired from IIS, we developed an automatic transition detection algorithm. This algorithm makes uses of cues like ring crossing detection, relative distance, person count, time constraint to segment the training sessions. In order to capture these observables cues from the input training session videos, we had to initially detect these individuals. For which, we utilized a pre-trained object detection model YOLOv7 [9]. It performed with a mAP of 0.9 for detecting the boxers on a confidence score of 0.5. The block diagram of the modified transition detection algorithm is shown in Fig. 1 (b). To determine whether a boxer is crosses the ring, we determined the centroid from the detected bounding box regions. Similarly, four virtual lines were placed on top of the boxing ropes along the sides. When the centroid co-ordinate crosses the virtual lines on any sides, we infer that a person has crossed the line (refer equation 1).

In the case of relative distance, Euclidean distance between the centroid of boxers is measured. When the distance falls below a threshold of 40, we infer that the boxers are in close proximity with each other (refer equation 2). The threshold of 40 was determined empirically across different bouts. Using equation 3, we determined the total number of individuals inside the ring. We performed this task by blacking other peripheral regions apart the ring. The time constraint was integrated to the above three cues, to make the approach more robust, i.e., a bout takes for a period of 120s followed by a resting period of 60s. The decision was taken based on majority rule. In other words, only when a minimal of two task's condition is satisfied, we say that the transition has occurred.

$$Ring\ crossing = \begin{cases} 1, & if\ \left(\frac{B_x - VL_x}{B_y - VL_y}\right) = 0 \\ 0, & if\ \left(\frac{B_x - VL_x}{B_y - VL_y}\right) \neq 0 \end{cases} \quad (1)$$

Here, $(B_x, B_y), (VL_x, VL_y)$ corresponds to boxers' and virutal lines' x and y coordinates

$$Close\ proximity = \begin{cases} 1, & if\ ED \leq 40 \\ 0, & if\ ED > 40 \end{cases} \quad (2)$$

Here, ED refers to the Euclidean distance

$$Person\ Count = \begin{cases} 1, & if\ DO = 3 \\ 0, & if\ 3 < DO > 3 \end{cases} \quad (3)$$

Here, DO refers to the total number of detected object in the ring

### C. In-bout Person Re-identification

The objective is to assign an ID to each individual in a bout and maintain it throughout the bout. To perform this task, we used two approaches namely (i) descriptor tracking; (ii) pose tracking, for evaluating their usability more complex data like combat sports.

*Descriptor tracking:* This approach is performed by utilizing a DeepSort [8] algorithm. In contrast to the conventional deepsort algorithm, we used YOLOv7 for detection. We normalized the positional and the appearance cost matrix to prevent the bias towards the larger factor. Euclidean distance was used to measure the distance across the cost matrix. Moreover, we kept the association cost ($\lambda = 0.8$) biased towards the positional cost matrix to make the descriptor tracking to be invariant to the resemblance of boxing attire. The track age criteria are kept as 10,000 frames, this helped to minimize faulty update of IDs.

*Pose tracking:* As poses of boxers vary rampantly, we segmented the bouts into multiple mini-bout segments of 120 frames. Then, we utilized a pre-trained Alphapose for landmark estimation. The estimated landmarks of each players get tracked through PoseFlow network [11]. For correlating the tracked information present in these mini-bouts (refer Fig. 1 (c)), we made use of Euclidean distance. The mean distance of the landmarks (two shoulder, two hip) points of all IDs from the last frame of the 1st mini-bout are calculated. Similarly, the mean distance of those landmarks of all IDs from the 1st frame of the 2nd mini-bout gets calculated. The IDs whose mean distance are minimal are assumed to be of the same individual (mini-bout integration module). This process is repeated and the tracking points of segmented mini-bouts are temporally correlated to continuous track.

## IV. RESULTS AND DISCUSSION

To assess the performance of proposed boxer, separate experiments were carried out on (i) transition detection; (ii) person re-identification separately. Experiment-1 deals with the performance of transition detection on these training videos. In this case, the total number of segmented bouts should be equal to that of the actual bouts that took place in a training session. The results obtained are shown in Fig. 2.

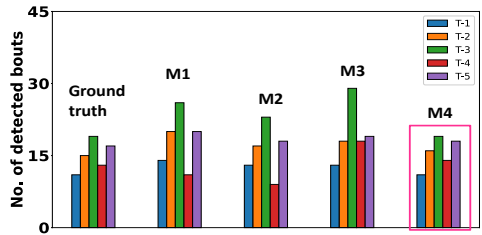

**Fig. 2** Transition detection algorithm
(T1 to T5 are the different training session. M1- person count, M2-ring crossing, M3-relative distance, M4-modified transition detection. The boxed region indicates that the proposed is nearer to the ground truth)

It can be seen that the modified algorithm over-segments bout segments in some instances. This may be because of the following reasons – (i) YOLOv7 failed to detect individuals at the time of transition resulting in individuals; (ii) boxers of the next bout enters the present bout within the time constraint; (iii) people moving along the side of the ring hangs onto the ropes resulting on line crossing. Experiment-2 deals with the person re-identification task. Descriptor tracking and pose tracking approaches were performed on these segmented bout videos. Here, the metrics like ID updation and ID switching should be ideally zero. The results obtained are shown in Fig. 3. To show the efficacy of the proposed methods, we have implemented the works of [8, 10, 11] on the collected training videos. It can be seen from the Fig. 4 (a) the ID updation becomes ideal (IDU = 0) in the case of the proposed pose tracking approach.

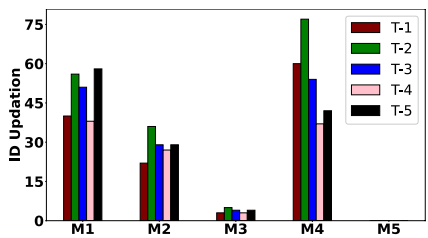

**Fig. 3** ID updation
(Here M1– Sort [11], M2 – DeepSort [8], M3 – Modified descriptor tracking, M4 – Alphapose+PoseFlow [10], M5- Modified pose tracking. Ground truth of IDU=0, and IDS=0)

While the descriptor tracking had the second least incorrect ID updates. Implementing Alphapose + PoseFlow on the actual bout video resulted in abrupt increase in the ID updation and ID switching. But, the proposed approach of integrating the estimated poses of Alphapose + Poseflow spatio-temporally across mini-bouts helped to minimize the updation and switching drastically (IDS =0). The result obtained clearly shows that (i) descriptor tracking (IDU~3; IDS~8); (ii) pose tracking (IDU=0; IDS=0) approaches introduced in this work are better than the existing SOTA (IDU>~10, IDS>~20) techniques. The reason for degradation in performance of descriptor tracking is due to merging complexity when the boxer comes too close (clinching) to each other. Pose tracking handles this issue through correlating the estimated landmarks generated. From this experiment we can infer that pose tracking on mini-bouts is ideal for dynamic sports like boxing.

## V. CONCLUSION

In this work, we introduce a vision-based system that relies on economic video capture system (single fixed viewpoint). Specifically, we propose modified algorithms for automatic transition detection system and person re-identification systems that shows significant improvements over state-of-the-art techniques. The modified automatic transition detection algorithm does not give ideal success rate. It has a success rate of 90% and we hypothesize that this occurred due to presence of more humans on the ring side. While the line-of-sight calculation degrades when the pose estimation network fails to detect the landmarks. We are working on these problems to further improve these algorithms.